\documentclass[letterpaper]{article} % DO NOT CHANGE THIS
\usepackage[preprint]{aaai2027}  % DO NOT CHANGE THIS
\usepackage[hyphens]{url}  % DO NOT CHANGE THIS
\usepackage{graphicx} % DO NOT CHANGE THIS
\usepackage{natbib}  % DO NOT CHANGE THIS AND DO NOT ADD ANY OPTIONS TO IT
\usepackage{caption} % DO NOT CHANGE THIS AND DO NOT ADD ANY OPTIONS TO IT
\usepackage{algorithm}
\usepackage{algorithmic}

\usepackage{newfloat}
\usepackage{listings}
\DeclareCaptionStyle{ruled}{labelfont=normalfont,labelsep=colon,strut=off} % DO NOT CHANGE THIS
\floatstyle{ruled}
\newfloat{listing}{tb}{lst}{}
\floatname{listing}{Listing}

\usepackage{booktabs}

\usepackage{amssymb}
\usepackage{textcomp}
\usepackage{xspace}
\usepackage{xcolor}
\usepackage{placeins}
\usepackage{amsmath}
\usepackage{multirow}
\usepackage[title]{appendix}

\usepackage{tikz}
\usetikzlibrary{arrows.meta, positioning, shapes.geometric}

\newcommand{\sys}{\textsc{Forethought}\xspace}

\newcommand{\mcite}[1]{{\color{purple}\textbf{[?]}}}
\newcommand{\tausq}{\ensuremath{\tau^2}}

\newcommand{\slmprim}[1]{\textsc{#1}\ensuremath{^{\dagger}}}
\newcommand{\symprim}[1]{\textsc{#1}}

\title{\sys: Verifiable Reasoning from Neurosymbolic Primitive Programming}
\author{
    Vishvesh Bhat\textsuperscript{\rm 1},
    Emmanuel Anaya Gonzalez\textsuperscript{\rm 1,\rm 2},
    Jay Vaghasiya\textsuperscript{\rm 1},
    Asad Aali\textsuperscript{\rm 3}
}
\affiliations{
    \textsuperscript{\rm 1}CoreThink AI

    \textsuperscript{\rm 2}UCSD

    \textsuperscript{\rm 3}Stanford University
}

\begin{document}

\maketitle

\begin{abstract}
Modern agentic workflows decompose user requests into sequences of tool calls
with correctly resolved parameters, the results of which are processed through
reasoning traces in the language model's context window. The prevailing route to
improving such reasoning is test-time scaling, which trains models to search over
long chains of thought; but the resulting capability is entangled in model
weights, cannot be checked step by step, and is costly at inference. We argue
instead that reasoning can be placed \emph{outside} the responding model's
weights, as an explicit program of small specialized models and symbolic checks
that can be inspected and checked. We present \sys, a
neurosymbolic reasoning system built from a library of symbolic and neural
primitives composed through a typed domain-specific language. The results are
\textit{reasoning programs}: concrete representations of the model's work, in
which every step carries an output contract that is checked at design time and
enforced at execution time. Symbolic steps hold by construction, while learned
steps are contract-checked and empirically validated, giving an explicit
boundary between what is guaranteed and what is measured.
Instantiated as a tool-calling execution kernel and evaluated on four standard
tool-calling benchmark suites, \sys improves end-to-end accuracy by a median of
32\% relative to prompting the same base model directly, and reaches the highest
aggregate accuracy among the evaluated configurations, though only by a margin
within single-run noise over a frontier model. Systems pairing small open models
with \sys's fixed small-model layer match that frontier model, and a pipeline
with a non-reasoning responder outperforms its reasoning sibling while requiring
only a single fine-tuning stage.
Removing the contract-and-control layer erases most of these gains, showing
that checked structure, not decomposition alone, drives the improvement.
Our results show the potential of rich symbolic orchestrators for building
language-model agents that are more accurate, reliable, and auditable.
\end{abstract}


% Introduction Section - Agentic Reasoner Paper
% Save as sections/intro.tex
% Citation keys: deepseek-r1, rlm-blueprint, gepa, dspy, severa, bfcl,
%   taubench, logiclm, proofofthought

\section{Introduction}
\label{sec:intro}

Large language models are increasingly deployed as agents that reason over
multiple steps and invoke external tools. The shift from single-turn
generation to multi-turn, tool-using agents makes \emph{reasoning
reliability} central: an agent must select the correct function, resolve its
parameters from context, respect dependencies, and maintain state across turns,
and one incorrect step can invalidate an entire trajectory~\citep{bfcl,taubench}.
As these systems enter production and regulated domains, reasoning must be not
only accurate but also efficient, cost-effective, and auditable.

Existing approaches improve reasoning in two broad ways. Test-time scaling
trains or prompts models to produce extended reasoning traces and search over
chains, trees, or graphs~\citep{deepseek-r1,besta-rlm}. Such capability is
costly and model-specific, and its stochastic trajectories cannot be checked
step by step.
Alternatively, declarative frameworks and prompt optimizers make multi-step
pipelines explicit and optimize their prompts or demonstrations~\citep{dspy,gepa},
but do not check intermediate semantic outputs. Agentic tool-calling lies in
the gap: it requires intermediate reasoning that is structured enough to
check, yet flexible enough to handle decisions such as intent coverage and
function matching, whose correctness cannot be decided by simple formal rules.

We present \sys, a neurosymbolic reasoning system for agentic tool-calling that
makes reasoning explicit, contract-checked, and model-agnostic. \sys{} decomposes
reasoning into a library of \emph{primitives}, narrow operations implemented
either symbolically or as fine-tuned small language models (SLMs), each with a typed
\emph{output contract}. A DSL composes primitives into reasoning programs, and
an execution engine checks every output against its contract, triggers recovery
when a check fails, and records a per-step trace. Because a program is an
analyzable data structure rather than an opaque chain, developers can check its
structural properties and review its empirically validated semantic operations
before deployment, instead of only discovering failures at runtime. The design
also draws an explicit \emph{assurance boundary}: properties enforced by
symbolic primitives hold by construction, while learned decisions are
contract-checked at runtime and validated empirically.

Our contributions are the following:
\begin{itemize}
    \item We introduce \sys, a neurosymbolic reasoning architecture for
    tool-calling built from contract-bearing symbolic and SLM-based
    primitives composed through an embedded, typed DSL with per-step checking
    and contract-triggered recovery.
    \item We formalize tool-calling as five composable sub-processes with
    explicit contracts and cross-process invariants, and give a design-time
    verification workflow that separates guaranteed structural properties from
    empirically validated semantic ones.
    \item We show that \sys{} improves end-to-end accuracy by a median of 32\%
    relative to prompting the same base model directly, and brings small
    responder models to frontier-level aggregate accuracy, with the highest
    aggregate score among the evaluated configurations (by a margin within
    single-run noise over the strongest frontier baseline).
    \item We show that removing the contract-and-control layer while keeping the
    primitives erases most of these gains, sometimes dropping accuracy below
    vanilla prompting, and that a pipeline with a non-reasoning responder
    outperforms its same-family reasoning model by 34 points while requiring only
    a single supervised fine-tuning stage on a 12B model.
\end{itemize}

% Related Work Section - Agentic Reasoner Paper
% Save as sections/rel_work.tex
% Citation keys used (add to refs.bib):
%   deepseek-r1, besta-rlm, gepa, dspy, severa, llmcompiler, chameleon,
%   bfcl, taubench, logiclm, proofofthought, nesypr

\section{Related Work}
\label{sec:related}

\paragraph{Reasoning Enhancement and Test-Time Scaling}
Reasoning language models such as DeepSeek-R1~\citep{deepseek-r1} develop
extended chain-of-thought, reflection, and self-verification behaviors through
large-scale reinforcement learning. Related work organizes inference-time
reasoning over chains, trees, and graphs, together with search strategies and
supervision schemes, into a common design space~\citep{besta-rlm}. These
methods can improve task performance, but their reasoning capability remains
entangled in model weights, their trajectories cannot be checked step by step,
and their training and inference costs can be substantial. We instead represent
reasoning as an explicit, reusable program with checked intermediate outputs.

\paragraph{Programmatic and Verifiable LM Systems}
DSPy~\citep{dspy} and GEPA~\citep{gepa} make LM pipelines explicit and optimize
prompts, demonstrations, or task-specific prompt programs. They share our
interest in structured orchestration, but optimize textual components without
typed contracts or checks on intermediate semantic outputs. Closest in
architecture, Chameleon~\citep{chameleon} composes heterogeneous neural and
symbolic tools into explicit programs, and LLMCompiler~\citep{llmcompiler}
plans dependency-aware, parallel function calls. \sys{} shares their
programmatic view but adds what neither provides: a typed output contract on
every step, deterministic guards and recovery triggered by contract violations,
and an explicit boundary between guaranteed and empirically validated
operations. Logic-LM and Proof-of-Thought delegate reasoning to symbolic
solvers~\citep{logiclm,proofofthought}, and SEVerA~\citep{severa} pairs
first-order-logic contracts with verified fallbacks; all target domains with
clean formal semantics. \sys{} instead targets tool-calling, whose key semantic
operations (intent coverage, function matching) are not fully formalizable: it
enforces structural correctness deterministically and uses empirically
validated learned primitives for semantic decisions, without requiring a
verified fallback for every operation.

% Methodology Section - Agentic Reasoner Paper
% Paste this into your Overleaf project

\section{Methodology}
\label{sec:methodology}

\begin{algorithm}[!ht]
\caption{Tool-Calling Execution Kernel, implemented as a reasoning program in our DSL. Primitives marked $^{\dagger}$ are SLM-based; all others are symbolic. The base LLM is outside this loop and receives the resulting trace as context.}
\label{alg:kernel_small}
\begin{algorithmic}[1]
\REQUIRE input $\iota$, config $\gamma$, registry $\Omega$, turn context $\tau$

\STATE \textbf{// \textsc{Decompose}}
\STATE $\mathit{turns} \leftarrow \slmprim{Turns}(\iota)$
\STATE $\mathit{config} \leftarrow \symprim{Extract}(\gamma)$; $\mathit{tools} \leftarrow \symprim{Register}(\Omega)$

\STATE \textbf{// \textsc{Init-Scratchpad}}
\STATE $S \leftarrow \{k \mapsto v\ \textbf{for}\ (k, v)\ \textbf{in}\ \mathit{config}\}$

\STATE \textbf{// \textsc{Get-Order}}
\STATE $G \leftarrow \slmprim{BuildDepGraph}(\mathit{turns})$
\STATE \textbf{if} $\symprim{CycleDetect}(G)$ \textbf{then} \textsc{halt} with error
\STATE $T \leftarrow \symprim{TopoSort}(G)$

\FORALL{$t \in T$}
    \STATE \textbf{// \textsc{Check-Prerequisites}}
    \FORALL{$p \in \symprim{GetPrereqs}(t)$}
        \STATE $r \leftarrow \symprim{Check}(p, S)$
        \IF{$r = \mathit{Absent}$}
            \STATE $v \leftarrow \symprim{Execute}(\slmprim{TaskToCheck}(p))$
            \STATE $S \leftarrow S \cup \{p \mapsto v\}$;\ \ $r \leftarrow \symprim{Check}(p, S)$
        \ENDIF
        \IF{$r = \neg\mathit{Met}$}
            \STATE $t' \leftarrow \slmprim{TaskToMeet}(p)$; \textbf{emit} $\mathit{Blocked}(t')$
            \STATE \textbf{skip} to the next task in $T$
        \ENDIF
    \ENDFOR

    \STATE \textbf{// \textsc{Match-Functions-and-Parameters}}
    \STATE $f \leftarrow \slmprim{MatchFunc}(t, \Omega)$; $M \leftarrow \symprim{Params}(f)$;\ \ $B \leftarrow \emptyset$
    % \IF{$f = \bot$}
    %     \STATE \textbf{report} \textsc{MissingFunction}; \textbf{ continue} \COMMENT{abandon $t$ (INV-R)}
    % \ENDIF
    % \STATE $M \leftarrow \textsc{Params}(f)$;\ \ $B \leftarrow \emptyset$
    \FORALL{$m \in M$}
        \STATE $v \leftarrow \symprim{Lookup}(m, \tau, S)$
        \IF{$v = \bot$}
            \STATE $v \leftarrow \symprim{Execute}(\slmprim{TaskToGet}(m))$
            \STATE $S \leftarrow S \cup \{m \mapsto v\}$
            % \STATE \textbf{if} $v = \bot$ \textbf{then report} \textsc{MissingParam}($m$); \textbf{ break}
        \ENDIF
        \STATE $B \leftarrow B \cup \{m \mapsto v\}$
    \ENDFOR
    \STATE \textbf{if} $\mathrm{dom}(B) = M$ \textbf{then} $\symprim{Execute}(f, B)$
\ENDFOR
\end{algorithmic}
\end{algorithm}

We now present the architecture of \sys,
which consists of three core components: (1)~a library of reasoning primitives with checkable output contracts, (2)~a domain-specific language (DSL) for composing primitives into structured reasoning programs, and (3)~an execution engine that interprets programs, enforces contracts, and records traces. We describe each component and their interactions below.

\subsection{Reasoning Primitive Library}
\label{sec:primitives}

At the foundation of our system lies a library of reasoning \textit{primitives},
each implementing a narrow, well-defined reasoning operation, and can be distinguished into two semantic
groups.
\textit{Symbolic primitives} have semantics expressible as a formal program,
which by construction guarantees correctness with respect to the specified
operation: type checking, format validation, constraint propagation, pattern
matching, and structural transformations. \textit{SLM-based primitives} are
fine-tuned small language models, each trained on a task with
natural-language semantics: entity extraction, relation classification, argument
decomposition, or logical entailment checking. We restrict each SLM to a single
narrow operation because multi-step tool-use benchmarks show that errors
compound over long, open-ended trajectories~\citep{bfcl, taubench}.
All primitives are
further designed to be \textit{narrowly scoped}, \textit{pure} (stateless and
side-effect-free, enabling parallel execution and deterministic replay),
\textit{composable}, and \textit{contract-checkable} through a typed interface.

We call the system \emph{neurosymbolic} in a specific sense: symbolic
primitives, whose semantics are given by code, and neural primitives, whose
semantics are learned, are composed under a single typed program whose control
flow is symbolic. Neural components make the judgments that resist
formalization, while symbolic components fix the structure, check every neural
output, and decide what happens next.

Every primitive carries an \emph{output contract}: an explicit specification of
properties its output must satisfy, spanning type constraints (a valid JSON
object conforming to a schema), value constraints (extracted entities must appear
in the source text), and semantic constraints (a decomposition must cover the
original query). Contracts are checked at execution time; a violation identifies
the failing primitive, its inputs, and the violated constraint.

We distinguish two levels of assurance. Deterministic structural contracts are
checked exhaustively and guarantee the specified properties. Semantic operations
implemented by SLMs are validated empirically: their contracts detect outputs
violating an expressible condition but do not prove that every contract-satisfying
output is correct. Our use of \emph{verification} therefore denotes contract
checking plus empirical validation, not end-to-end formal correctness. Making
this boundary explicit, per primitive, is itself a design goal: it tells a
developer exactly which properties of a program are guaranteed and which must
be measured.

\subsection{Embedded Domain-Specific Language}
\label{sec:dsl}

% Using an embedded DSL enables several analyses that are impossible with generic Python code: 
% \textit{1)} type‑checked composition ensures that mismatched primitive interfaces are caught at program construction time rather than at runtime; 
% \textit{2)} it allows the execution engine to traverse the program tree and predict the number of LLM calls, expected latency, and inference cost before any computation occurs; 
% \textit{3)} correctness reasoning follows from the contracts of individual primitives and their valid compositions as checked via the type system
% and \textit{4)} program optimization becomes feasible, as redundant primitive calls can be eliminated, independent branches can be parallelized, and frequently co‑occurring patterns can be cached by transforming the AST prior to execution.

The orchestration of our primitives is done via a typed, imperative domain-specific language
embedded in Python. Compared with plain Python, its richer type system catches many
errors during program construction, so structural correctness follows from valid
composition of subprograms, and its explicit program structure makes caching,
parallelization, and static estimates of model-call counts possible.

The DSL provides basic composition operators that express common reasoning patterns:
\textit{Sequence} executes primitives in order, passing outputs forward.
\textit{Parallel} processes independent primitives concurrently.
\textit{Conditional} enables branch execution based on intermediate results or contract outcomes.
\textit{Iterate} repeats a subprogram until a termination condition holds or a fixed retry budget $k$ is exhausted.
\textit{Fallback} attempts a finite list of alternatives upon primitive failure or contract violation.

% A principal benefit of our architecture is that it permits system‑level reliability to be inferred from measurements at the primitive level.
Because every composition is bounded, each program halts. Because every step is checked, an error in one primitive is caught where it occurs rather than propagating silently: the engine retries or substitutes the failed primitive, and traces identify the least reliable primitives for retraining.
% Let $r_i$ denote the empirically measured reliability of primitive $i$—the probability that its output satisfies its contract on a randomly drawn input from the expected distribution.  
% For a sequential composition of $n$ primitives, a naïve bound would be $\prod_{i=1}^{n} r_i$, which deteriorates quickly as the chain lengthens.  

\subsection{Execution Engine}
\label{sec:execution}

The execution engine interprets reasoning programs, manages primitive
invocations, enforces contracts, and produces structured traces. 
% It traverses
% the program AST and evaluates each primitive by type: deterministic primitives
% run directly, while SLM-based primitives are dispatched to the appropriate
% fine-tuned model, with input formatting, output parsing, and retry logic handled
% by the engine rather than the user.
% \paragraph{Per-Step Verification.}
For each primitive's output, the engine checks it against the
existing contract. 
A satisfied contract lets execution proceed; 
a violation
invokes the program's error-handling strategy: retrying with modified inputs,
invoking a fallback, or halting with a structured error report.

% \paragraph{Symbolic Trace Generation.}
The engine records a structured trace for every invocation: primitive identity,
inputs, outputs, contract-check results, execution time, and (for SLM-based
primitives) model identity.
Traces localize failures for debugging, provide an
inspectable record for audit in regulated settings, and surface high-failure-rate
primitives for targeted retraining, without user instrumentation.
Runtime optimizations are described in Appendix~\ref{app:design-details}.

\subsection{Design-Time Verification}
\label{sec:design-time}

A distinguishing property of our architecture is that many properties of reasoning
programs can be checked, manually or automatically, before deployment rather
than surfacing at runtime. As the developer composes primitives, the system
validates type compatibility and well-formedness across interfaces; the program is
then run on representative inputs with full trace logging to confirm that each
primitive behaves as expected and that contract violations are caught. Confidence in the
whole program thus comes from checking its parts, mirroring work on design-time
verification of AI systems~\citep{decidable, buildingblocks}.

\section{\sys for Agentic Tool-Calling}
\label{sec:tool-calling}

We instantiate the general framework into the tool-calling domain,
targeting the agentic setting where a model must decompose user requests into
sequences of API calls with correctly resolved parameters. 
This concretizes the primitive library, composition operators, and contracts for a specific domain
while preserving the general architecture.

\paragraph{Formal Domain.}
The kernel operates over base sets $\mathbb{V}$ (values), $\mathbb{K}$ (key
names, a unified namespace for configuration, prerequisites, and parameters),
$\mathbb{T}$ (unit task descriptors), $\mathbb{F}$ (function identifiers),
$\mathbb{M} \subseteq \mathbb{K}$ (parameter names), and $\mathit{Sig}$ (function
signatures). It receives a raw multi-turn input $\iota$, an initial configuration
$\gamma$, and a tool registry $\Omega$.

\paragraph{Sub-Process Architecture.}
The kernel comprises five composable sub-processes, each implemented from
deterministic primitives, SLM-based primitives, or a combination. This
classification determines the verification strategy at each stage and makes
explicit where neural inference is required and where symbolic computation
suffices. SLM-based primitives are fine-tuned per operation with output
contracts as noted. Algorithm~\ref{alg:kernel_small} gives the main control
flow; Algorithm~\ref{alg:kernel} in Appendix~\ref{app:algorithm} is the same
kernel with its error-reporting branches and invariant annotations.

\begin{enumerate}
    \item \textsc{Decompose}$(\iota, \gamma, \Omega)$ \emph{(hybrid)}: separates
    raw inputs into an ordered task list $\mathit{turns}$, a configuration
    $\mathit{config}: \mathbb{K} \rightharpoonup \mathbb{V}$, and a registry
    $\mathit{tools}: \mathbb{F} \rightharpoonup \mathit{Sig}$. Configuration
    extraction and tool registration are deterministic schema parses; task
    derivation ($\mathit{Turns}$) is SLM-based, with a contract requiring every
    derived task to map to a function in $\Omega$ and the tasks to jointly cover
    the user's intent.

    \item \textsc{Init-Scratchpad}$(\mathit{config})$ \emph{(deterministic)}:
    seeds a mutable store $S: \mathbb{K} \rightharpoonup \mathbb{V}$ from
    $\mathit{config}$. $S$ persists for the run and grows monotonically: keys are
    never deleted, though values may be overwritten. Correctness is immediate:
    $\mathrm{dom}(S) = \mathrm{dom}(\mathit{config})$ and $S(k) = \mathit{config}(k)$.

    \item \textsc{Get-Order}$(\mathit{turns})$ \emph{(hybrid)}: builds a
    dependency graph $G=(V,E)$, where $(t_i, t_j) \in E$ means $t_i$ depends on
    $t_j$, detects cycles, and returns a topological order $T$ in which every
    task follows its dependencies: $\forall (t_i, t_j) \in E:\ \mathit{pos}(t_j) < \mathit{pos}(t_i)$.
    Graph construction is SLM-based, contracted so that a declared dependency
    corresponds to an actual data-flow relationship; cycle detection and
    topological sort are deterministic and correct by construction.

    \item \textsc{Check-Prerequisites}$(t, S)$ \emph{(mostly deterministic)}: for
    each prerequisite $p$, evaluates
    {\small
    \begin{equation}
    \text{Check}(p, S)\!=\!\begin{cases}
    \mathit{Met}       & p \!\in\! \text{dom}(S) \,\wedge\, \mathit{Sat}(p, S(p)) \\
    \neg\mathit{Met}   & p \!\in\! \text{dom}(S) \,\wedge\, \neg\mathit{Sat}(p, S(p)) \\
    \mathit{Absent}    & p \!\notin\! \text{dom}(S).
    \end{cases}
    \end{equation}
    }
    The membership check and branching are symbolic; the $\mathit{Sat}$
    predicate may be symbolic (type/non-null) or SLM-based (a semantic
    condition, returning a thresholded decision). $\mathit{Absent}$ triggers
    inline retrieval, writes the result to $S$, and re-runs \textsc{Check} on the
    retrieved value; since $p$ is now in $\mathrm{dom}(S)$, each key fires
    $\mathit{Absent}$ at most once. $\neg\mathit{Met}$ emits a blocking
    remediation task and skips the current task. Remediation and retrieval tasks
    are SLM-generated and yield a value in the required key's domain.

    \item \textsc{Match-Functions-and-Parameters}$(t, S, \tau, \Omega)$
    \emph{(hybrid)}: matches a function $f$ to $t$, resolves parameters $M$ by
    prioritized lookup (turn context $\tau$ before scratchpad $S$), and executes
    only when fully bound ($\mathrm{dom}(B) = M$). Function matching is SLM-based,
    contracted for signature compatibility; the lookup and the execution guard
    are deterministic. Missing values are retrieved via the same SLM-based pattern
    as above.
\end{enumerate}

\paragraph{Verification Strategy.}
This classification determines how each stage is verified. Symbolic
primitives, being deterministic, are checked exhaustively, whereas
SLM-based primitives are validated empirically against held-out test sets, with
contracts providing runtime checks that catch violations. Hybrid sub-processes
are assessed compositionally: contracts at the boundaries prevent detected
violations from propagating from SLM-based into deterministic components. This
enables the design-time workflow below but does not constitute a formal proof of
end-to-end semantic correctness.

\paragraph{Contracts and Invariants.}
Each sub-process enforces post-conditions as output contracts, and the following
invariants hold globally, enforced by the DSL's typed composition and the
engine's guards; a program violating any is rejected at construction or caught at
the first violating step:
\begin{itemize}
    \item \emph{Scratchpad monotonicity (INV-S):} $\mathrm{dom}(S_i) \subseteq \mathrm{dom}(S_j)$ for steps $i<j$.
    \item \emph{Termination (INV-T):} every run halts, either after the last task in $T$ or with an error when $G$ is cyclic, since $T$ is finite, each key fires $\mathit{Absent}$ at most once, and Iterate and Fallback are bounded.
    \item \emph{Execution precondition (INV-E):} $\textsc{Execute}(f, B)$ fires iff $\mathrm{dom}(B) = \mathit{Params}(f)$; no partial call is ever issued.
    \item \emph{Lookup priority (INV-L):} $\tau$ is always consulted before $S$.
    \item \emph{Non-fatal errors (INV-R):} an unmet prerequisite, missing function, or missing parameter abandons only the current task; execution continues with the next task in $T$.
\end{itemize}

\paragraph{Design-Time Verification on BFCL Tool-Calling.}
We illustrate the design-time verification methodology through our development of
the tool-calling execution kernel. Prior to deployment, we constructed a synthetic
validation corpus of 300 multi-turn scenarios, authored by us following BFCL's
multi-turn, deterministic evaluation methodology and not taken from any
evaluation set, covering the interaction patterns the kernel must
handle: cross-turn state carry-over, multi-step dependency chains, prerequisite
resolution, and parameter grounding across turns. For each, we manually worked
through the complete DSL execution (decomposition, scratchpad initialization, task
ordering, prerequisite checking, and function/parameter matching), producing an
annotated trace recording every sub-process decision and its justification. As a
representative example, Figure~\ref{fig:cross-turn} shows the Turn~2 trace for a
scenario in which a prior turn moved a file and the current turn searches within
it, illustrating how cross-turn state is managed through the scratchpad; the
resulting tool-call sequence matches the reference call sequence.

\begin{figure*}[t]
\centering
\footnotesize
\fbox{\parbox{0.97\textwidth}{
\textbf{Scenario:} file-system example, Turn 2 \quad|\quad
\textbf{Input ($\iota$):} ``Search the moved report for any mention of \texttt{revenue}.'' \\
\textbf{Scratchpad from Turn 1:} $S = \{\texttt{archive} \mapsto \mathit{created},\ \texttt{folder} \mapsto \texttt{'archive'},\ \texttt{file\_name} \mapsto \texttt{'report.txt'}\}$ \\[3pt]
\rule{\linewidth}{0.4pt}\\[3pt]
\begin{minipage}[t]{0.48\linewidth}
\textbf{1. \textsc{Decompose}:} \\
\quad $t_1$: navigate to \texttt{archive} \\
\quad $t_2$: search \texttt{report.txt} for pattern \\[3pt]
\textbf{2. \textsc{Get-Order}:} $t_1 \prec t_2$ \\
\quad {\scriptsize(enter directory before searching)} \\[3pt]
\textbf{3. \textsc{Check-Prereqs}($t_1$):} \\
\quad \textsc{Check}(\texttt{archive}, $S$) $= \mathit{Met}$ \\
\quad {\scriptsize(from Turn 1)} $\Rightarrow \langle \mathit{Ready}, S \rangle$
\end{minipage}%
\hfill
\begin{minipage}[t]{0.48\linewidth}
\textbf{4. \textsc{Match}($t_1$):} $f=\texttt{cd}$ \\
\quad \textsc{Lookup}(\texttt{folder}): $S \rightarrow \texttt{'archive'}$ \\[3pt]
\textbf{5. \textsc{Match}($t_2$):} $f=\texttt{grep}$ \\
\quad \textsc{Lookup}(\texttt{file\_name}): $S \rightarrow \texttt{'report.txt'}$ \\
\quad \textsc{Lookup}(\texttt{pattern}): $\tau \rightarrow \texttt{'revenue'}$ \\
\quad {\scriptsize(from user input, not $S$)}
\end{minipage}
\\[3pt]
\rule{\linewidth}{0.4pt}\\[3pt]
\textbf{DSL output:} \texttt{cd(folder='archive')}; \texttt{grep(file\_name='report.txt', pattern='revenue')} \\
\textbf{Result:} $\checkmark$ exact match with reference \quad|\quad functional correctness $\checkmark$ \quad|\quad DSL compliance $\checkmark$
}}
\caption{Cross-turn state management. Turn~2 resolves the directory and file from
scratchpad state established by Turn~1, while the search pattern is resolved from
the current turn's input ($\tau$). The scratchpad makes cross-turn dependencies
explicit and verifiable, whereas unstructured approaches rely on implicit
context-window management.}
\label{fig:cross-turn}
\end{figure*}

%% We removed this example for now
% A more complex example, shown in Figure~\ref{fig:cross-task}, involves cross-task parameter resolution within a single turn. When a user requests a passenger name correction on a reservation, the DSL decomposes this into two ordered tasks: retrieve reservation details (to obtain the existing passenger's date of birth), then update the passenger record with the corrected name while preserving the unchanged DOB. The parameter resolution for the update task demonstrates the lookup priority: the new name comes from the turn context $\tau$ (user-provided), while the DOB comes from the scratchpad $S$ (retrieved by the prior task). This cross-task state management is made explicit and verifiable by the scratchpad formalism, whereas in unstructured approaches it would be implicit in the model's context window.

\paragraph{Iterative DSL Refinement.}
The 300 worked solutions were also the primary mechanism for refining the DSL: each one that revealed friction or failure in a sub-process specification led to a concrete revision, such as the $\mathit{Absent}$ branch, the prioritized lookup, and the monotonicity invariant (Appendix~\ref{app:design-details}), rather than tuning post hoc to benchmark outputs.

\section{Experiments}
\label{sec:experiments}

We evaluate \sys on four tool-calling benchmark suites spanning multiple domains,
comparing it against agentic reasoning frameworks and vanilla LLM prompting.
We describe the fine-tuning procedure for our SLM-based primitives, report main results,
ablate the contract-and-control layer, and compare against test-time scaling.

\paragraph{System configuration.}
In every \sys{} row, the five fine-tuned Gemma-12B primitives (Section~\ref{sec:finetuning}) produce the same structured reasoning trace; the row's base model receives it in context and generates the final response from it.
We attribute differences between rows to how well each base model interprets the trace. Vanilla rows use the base model alone.

\subsection{Benchmarks}
\label{sec:benchmarks}

The evaluation suite comprises four benchmarks: \textit{\tausq-Bench}~\citep{taubench,tau2bench} (Airline, Retail, and Telecom domains), \textit{BFCL v4 Multi-Turn-Base}~\citep{bfcl}, \textit{LiveMCPBench}~\citep{livemcpbench},
and \textit{BFCL v3 Multi-Turn-Base}~\citep{bfcl}.
Together they span distinct function registries, tool-discovery settings, and multi-turn interactions in which later turns depend on state established earlier, directly testing our scratchpad-based cross-turn state management.

A recent audit of BFCL v4~\citep{benchmark-audit}
found that the standard scoring pipeline exhibited several inaccuracies:
some ground‑truth annotations were erroneous, and functionally equivalent outputs were mistakenly penalized.
Three expert annotators (about 89 annotator-hours in total) manually inspected failing cases and corrected these scoring errors. We report both the original automated scores and the human‑corrected scores, but the corrected labels have not been independently audited and we do not claim the annotators were independent of system development (see Appendix~\ref{app:bfcl-corrections}).

\subsection{Baselines}
\label{sec:baselines}

We compare \sys against three distinct paradigms for tool-calling across standard open-weights and proprietary foundation models. 
First, we evaluate \textit{vanilla LLM prompting}, where models receive function signatures, turn history, and user requests directly in prompt context. 
This establishes baseline performance for unstructured prompting across a broad spectrum of open and proprietary models.
Second, we evaluate \textit{Recursive Language Models (RLM)}~\citep{rlm-blueprint}, an agentic framework for long-horizon tasks that enables base language models to dynamically execute Python code and inspect intermediate environment state. 
Third, we compare against \textit{Genetic Evolution Prompt Optimization (GEPA)}~\citep{gepa}, a prompt compilation framework that uses reflection models and evolutionary search to discover domain-optimized prompts.
Setups for baselines are further detailed in Appendix~\ref{app:eval-details}.

% For both agentic baselines (RLM and GEPA), we perform hyperparameter sweeps across execution budgets, iteration limits, and reflection models, reporting their best-performing configurations on each benchmark to ensure comparison against fully optimized baselines. 
% Complete sweep details and hyperparameter setups are provided in Appendix~\ref{app:eval-details}.

\subsection{SLM Fine-Tuning}
\label{sec:finetuning}

Each neural primitive in our DSL is implemented as a separately fine-tuned small language model using Gemma 12B~\citep{gemma} as the base architecture. 
% Training examples were derived from 300 manually verified DSL solutions and augmented with synthetic scenario variations, yielding 16{,}502 total examples across the five primitives (Table~\ref{tab:finetuning-data}). 
Training examples were derived from 300 manually verified DSL solutions and augmented with synthetic scenario variations, yielding 16{,}502 total examples across the five primitives. 
Models were fine-tuned for 4 epochs ($lr = 1 \times 10^{-5}$) and evaluated using strict exact-match on structured outputs.

Each fine-tuned primitive is validated independently against its output contract before integration into the full system. 
For example, validation for \textsc{Get-Order} checks that the produced ordering respects all declared dependencies. 
Primitives that fail to meet reliability thresholds on validation are retrained with additional data before deployment.
Fine-tuning significantly improves per-primitive exact-match accuracy on held-out synthetic test sets, with the weighted average rising from 76.4\% to 92.5\%. 
The largest gains occur on the semantically complex tasks, \textsc{Decompose} (+23.1\%) and \textsc{Check-Prerequisites} (+14.5\%). 
Dataset construction details and per-primitive examples are provided in Appendix~\ref{app:finetuning-examples}.

\subsection{Results}
\label{sec:results}

% Main Results Table - \sys{} paper
% Spans both columns. Requires booktabs (already loaded). MCP-Atlas excluded (v1).
% Bold = best in column across all methods. Best Vanilla per column is underlined.

\begin{table*}[t]
\centering
\small
\caption{Main results across benchmarks.
$\tau^2$-Bench domains (Airline/Retail/Telecom), BFCL v4 and v3 multi-turn-base, LiveMCPBench,
and BFCL v4 with human review (v4+H). Avg$_1$ excludes v4+H; Avg$_2$ uses v4+H in
place of v4. Best per column in \textbf{bold}. All scores are from a single run.
Entries marked $^\ast$ errored out and are counted as 0 in the averages.}
\label{tab:main-results}
\setlength{\tabcolsep}{4pt}
\begin{tabular}{@{}ll rrr rr r r rr@{}}
\toprule
& & \multicolumn{3}{c}{\textbf{$\tau^2$}} & \multicolumn{2}{c}{\textbf{BFCL}} & & & & \\
\cmidrule(lr){3-5}\cmidrule(lr){6-7}
\textbf{Method} & \textbf{Model} & \textbf{Air.} & \textbf{Ret.} & \textbf{Tel.}
& \textbf{v4} & \textbf{v3} & \textbf{LMCP} & \textbf{v4+H} & \textbf{Avg$_1$} & \textbf{Avg$_2$} \\
\midrule
\multirow{6}{*}{\shortstack[l]{\textbf{\sys{}}\\(ours)}}
 & haiku 4.5      & 82.0 & 92.1 & 98.0 & 73.0 & 41.5 & 69.0 & \textbf{91.5} & 75.9 & 79.0 \\
 & kimi-k2.6      & 82.0 & \textbf{93.0} & 98.0 & 70.5 & 42.5 & 74.7 & 89.0 & \textbf{76.8} & \textbf{79.9} \\
 & mistral 3.5    & 74.0 & 91.2 & 96.0 & 70.5 & 43.5 & 75.0 & 85.0 & 75.0 & 77.5 \\
 & gpt-oss-120b   & 84.0 & 85.1 & 91.0 & 64.0 & 36.0 & 76.0 & 90.0 & 72.7 & 77.0 \\
 & deepseek-v3    & 84.0 & 79.8 & 86.8 & 70.5 & 39.5 & 73.7 & 87.5 & 72.4 & 75.2 \\
 & gpt-5.4-mini   & \textbf{88.0} & 77.2 & 86.0 & 67.0 & 44.0 & 75.0 & 85.5 & 72.9 & 76.0 \\
\midrule
\multirow{8}{*}{Vanilla}
 & opus-4.7       & 80.0 & 87.7 & \textbf{98.2} & \textbf{77.0} & 42.0 & 75.0 & 85.5 & 76.7 & 78.1 \\
 & sonnet-4.6     & 80.0 & 77.2 & 97.0 & 73.0 & \textbf{45.0} & 74.7 & 83.5 & 74.5 & 76.2 \\
 & haiku-4.5      & 52.0 & 75.4 & 53.5 & 62.0 & 42.0 & 70.5 & 73.0 & 59.2 & 61.1 \\
 & kimi-k2.6      & 80.0 & 78.0 & 95.0 & 70.0 & 39.0 & 72.0 & 75.0 & 72.3 & 73.2 \\
 & gpt-oss-120b   & 52.0 & 64.9 & 46.5 & 61.0 & 30.0 & 53.0 & 69.5 & 51.2 & 52.7 \\
 & deepseek-v3    & 64.0 & 78.0 & 57.9 & 46.5 & 28.0 & 54.0 & 61.5 & 54.7 & 57.2 \\
 & deepseek-r1    & 48.0 & 44.7 & 7.0  & 57.0 & 30.5 & 43.0 & 75.5 & 38.4 & 41.5 \\
 & gpt-5.4-mini   & 36.0 & 50.0 & 21.1 & 38.5 & 22.0 & 42.0 & 56.0 & 34.9 & 37.9 \\
\midrule
\multirow{3}{*}{RLM}
 & kimi-k2.6      & 86.6 & 80.6 & 79.6 & 56.0 & 29.0 & 14.7 & 68.0 & 57.8 & 59.8 \\
 & gpt-oss-120b   & 21.2 & 34.6 & 54.8 & 15.5 & 1.0  & 5.3  & 28.0 & 22.1 & 24.2 \\
 & deepseek-v3    & 0.0$^\ast$ & 2.5 & 0.0$^\ast$ & 2.5 & 1.5 & 0.0$^\ast$ & 5.5 & 1.1 & 1.6 \\
\midrule
\multirow{3}{*}{GEPA}
 & kimi-k2.6      & 80.0 & 78.1 & 81.0 & 70.0 & 41.0 & \textbf{77.0} & 87.5 & 71.2 & 74.1 \\
 & deepseek-v3    & 36.0 & 51.8 & 40.4 & 26.5 & 19.5 & 40.0 & 41.5 & 35.7 & 38.2 \\
 & gpt-oss-120b   & 60.0 & 62.0 & 66.0 & 57.5 & 31.0 & 63.2 & 74.0 & 56.6 & 59.4 \\
\bottomrule
\end{tabular}

\vspace{2pt}
{\footnotesize $^\ast$RLM requires the base model to write code to inspect the input; in these instances the generated code errored out.}
\end{table*}
% Table~\ref{tab:main-results} presents our main results across all benchmarks and base models. 
% We report scores for our neurosymbolic reasoning engine (\sys{}) with multiple inference models, vanilla LLM prompting baselines, and the best configurations of RLM and GEPA from the sweeps reported in Tables~\ref{tab:rlm-sweep} and~\ref{tab:gepa-sweep}. 
% Several observations emerge from these results.

Table~\ref{tab:main-results} presents our main results across all benchmarks and base models; all scores reflect a single run per configuration.
% \paragraph{\sys{} consistently improves base model performance.}
For every base model evaluated under both \sys{} and vanilla prompting,
\sys{} produces a higher aggregate average, with a median relative gain of
32\% in Avg$_1$ across the five matched pairs.
The improvement is largest for weaker models: Haiku~4.5 rises from 52.0 to 82.0 on $\tau^2$ Airline and from 53.5 to 98.0 on $\tau^2$ Telecom.
Gains are not uniform at the level of individual benchmarks: Haiku~4.5 with \sys is slightly below vanilla on LiveMCPBench (69.0 vs.\ 70.5) and BFCL v3 (41.5 vs.\ 42.0), though its aggregate is 16.7 points higher.
Stronger models also improve but by smaller margins, consistent with weaker models having more to gain from an externally supplied, checked trace.
\sys{} on kimi-k2.6 reaches 76.8 Avg$_1$, on par with vanilla Opus~4.7 (76.7), a much larger frontier model; with single runs we read this as parity rather than an advantage. \sys is best on only two of the six automated columns (Airline and Retail), and differences of about one point are within run-to-run noise.

\sys also compares favorably to other language-model harnesses. RLM's best scores are competitive on individual benchmarks (86.6 on Airline with kimi-k2.6) but carry error rates of 10--34\% that lower effective accuracy on full denominators; \sys exceeds its pass rates on every benchmark except Airline (82.0 vs.\ 86.6 with the same model), without this penalty. GEPA attains the best individual LiveMCPBench score, but its results vary widely across models and benchmarks, whereas \sys achieves higher aggregate averages than every GEPA configuration in Table~\ref{tab:main-results} with no per-benchmark tuning.
%
% \paragraph{Human-reviewed scoring narrows the gap with ground truth.}
Human review of BFCL v4 raises the scores of all systems, vanilla included, and raises \sys{} scores by 14.5--26 points (e.g., Haiku~4.5 from 73.0 to 91.5).

We treat the automated scores (Avg$_1$) as primary, and our claims \emph{do not depend on human review}: under the original benchmark scoring, \sys lifts systems built on small and inexpensive base models to near frontier-level aggregate scores (Haiku~4.5 and Mistral~3.5 reach 75.9 and 75.0, against 76.7 for vanilla Opus~4.7).

% Ablation Studies section - Forethought paper
% Insert into sections/eval.tex (replaces the placeholder ablation subsection).
% Requires booktabs. Table is single-column.

\subsection{Ablation: Contractual Verification and Control}
\label{sec:ablations}

Our central claim is that the \emph{contractual verification-and-control layer} (CVC),
the bundle of output contract definitions, enforcement mechanisms, prerequisite checking, and the associated recovery paths,
is a major driver of the accuracy of a neurosymbolic program in our system,
beyond what decomposition into primitives provides on its own.
If this is true, removing the CVC while keeping the same primitives and program structure should sharply degrade performance.
We compare three settings with the same base model:
\textit{1)} vanilla direct prompting of the base model,
\textit{2)} \sys without the CVC, in which the same primitives are chained directly, each consuming the preceding output without validation,
and \textit{3)} the complete \sys system.

% \paragraph{Configurations.}
% We compare three settings on the same base model:
% \begin{itemize}
%     \item \textbf{Vanilla:} direct prompting of the base model, with no
%     neurosymbolic program.
%     \item \textbf{\sys{} without contractual verification:} the same primitives
%     composed into a program, but with output contracts, contract enforcement,
%     prerequisite checking, and the associated recovery paths removed. Primitives
%     are chained directly, each consuming the preceding primitive's output without
%     validation.
%     \item \textbf{\sys{} (full):} the complete system, with contracts enforced at
%     every step and violations handled through the mechanisms of
%     Section~\ref{sec:tool-calling}.
% \end{itemize}

\begin{table}[t]
\centering
\footnotesize
% \caption{Ablating the contractual verification-and-control layer (CVC). v4+H is BFCL v4 under human
% review. Avg is over $\tau^2$ Airline, $\tau^2$ Retail, and BFCL v4;
% Avg$_H$ substitutes the human-reviewed score. Removing the bundled layer costs 38.6 points on
% GPT-OSS-120B and 16.0 on Haiku~4.5.}
\caption{Accuracy after ablating the contractual verification-and-control layer (CVC). Avg is over Airline, Retail, and BFCL v4; Avg$_H$ uses the human-reviewed v4 score. Single runs.}
\label{tab:ablation}
\setlength{\tabcolsep}{2pt}
\begin{tabular}{@{}l cc cc cc@{}}
\toprule
& \multicolumn{2}{c}{\textbf{$\tau^2$}} & \multicolumn{2}{c}{\textbf{BFCL}} & & \\
\cmidrule(lr){2-3}\cmidrule(lr){4-5}
\textbf{Config.} & \textbf{Air.} & \textbf{Ret.} & \textbf{v4} & \textbf{+H} & \textbf{Avg} & \textbf{Avg$_H$} \\
\midrule
\multicolumn{7}{@{}l}{\textit{GPT-OSS-120B}} \\
Vanilla        & 52.0 & 64.9 & 61.0 & 69.5 & 59.3 & 62.1 \\
\sys{} w/o CVC & 47.4 & 56.0 & 14.0 & 18.0 & 39.1 & 40.5 \\
\sys{} (full)  & \textbf{84.0} & \textbf{85.1} & \textbf{64.0} & \textbf{90.0} & \textbf{77.7} & \textbf{86.4} \\
\midrule
\multicolumn{7}{@{}l}{\textit{Claude Haiku 4.5}} \\
Vanilla        & 52.0 & 75.4 & 62.0 & 73.0 & 63.1 & 66.8 \\
\sys{} w/o CVC & 60.0 & 69.3 & 70.0 & 74.5 & 66.4 & 67.9 \\
\sys{} (full)  & \textbf{82.0} & \textbf{92.1} & \textbf{73.0} & \textbf{91.5} & \textbf{82.4} & \textbf{88.5} \\
\bottomrule
\end{tabular}
\end{table}

% \paragraph{The verification-and-control layer is necessary for the observed gains.}
% Table~\ref{tab:ablation} shows that removing the bundled layer sharply reduces
% performance. 
On GPT-OSS-120B the average falls from 77.7 to 39.1, a loss of 38.6
points, roughly $2\times$ the system's entire gain over vanilla prompting (18.4
points). 
On Haiku~4.5 the loss is 16.0 points, against a full-system gain over
vanilla of 19.3. 
These numbers indicate that the full-system gains cannot be explained by the unchecked composition of primitives: the CVC plays a fundamental role.
On GPT-OSS-120B the ablated configuration scores 39.1, \emph{below} vanilla prompting at 59.3: unchecked decomposition creates more places to fail, and errors propagate silently into later steps.
Because the CVC is removed as a whole, this ablation establishes the value of the checked control layer but does not apportion credit among its components.

% Because the learned primitives and task decomposition are retained
% in the ablated configuration, these results show that 
% CVC is necessary to explain the full-system gains.

% \paragraph{Without the verification-and-control layer, a neurosymbolic program can be worse than no program at all.}
% prompting at 57.6. 

% This is expected rather than anomalous: in an unverified
% chain, an error at any step propagates silently into every subsequent step, so
% the probability of a correct final trajectory decays multiplicatively with
% program length. Decomposition alone therefore does not help---it creates more
% places to fail. What makes decomposition profitable here is the combined ability
% to check each step, enforce prerequisites, and recover from detected failures.
% The structure is therefore beneficial in these experiments only when paired with
% the full verification-and-control layer.

% Additionally, the gain from the full layer is far larger for GPT-OSS-120B ($+38.6$) than for Haiku~4.5 ($+16.0$). 

% \paragraph{The observed benefit is larger for the higher-error base model.}
The CVC contributes more on GPT-OSS-120B (38.6 points) than on Haiku~4.5 (16.0); the main results (Section~\ref{sec:results}) show a similar pattern, and we do not separate whether it reflects how each model handles an unchecked trace or other factors.

% \paragraph{Scope.}
% This ablation covers two base models and removes the verification-and-control
% layer as a single bundle: output contracts, enforcement, prerequisite checking,
% and recovery paths. It does not identify how much of the observed difference is
% caused by any individual component. Separating these effects would require
% additional controlled configurations that we have not run. The magnitude of the drop on
% GPT-OSS-120B is larger than the error-propagation account alone would predict;
% we do not have a controlled explanation for this difference and leave it to
% future work. We likewise do not claim the effect size generalizes beyond the
% models and benchmarks evaluated here. The consistent direction of the effect
% across two models of differing capability, and its consistency with the
% error-rate mechanism above, is what we take as support for the claim.

\subsection{Comparison with Test-Time Scaling}
\label{sec:tte}

% A central question for the field is whether reasoning improvements should come from scaling inference compute (test-time scaling) or from structured reasoning programs. Test-time scaling approaches---exemplified by DeepSeek-R1, OpenAI's o-series, and similar reasoning models---invest additional compute at inference to search over reasoning chains, producing longer and more deliberate outputs. These approaches have demonstrated strong results but at significant cost: reasoning models consume 5--50$\times$ more tokens per query than their non-reasoning counterparts, and the reasoning capability is baked into a specific model through specialized training rather than being transferable across models.
A central question for our work is how our reasoning programs compare to scaling compute at inference. 
% Reasoning models have demonstrated strong results in logic and math but at significant inference compute cost.
To evaluate this, we pick DeepSeek-V3 \cite{deepseek-v3} and its reasoning successor DeepSeek-R1 \cite{deepseek-r1} and compare them in the following settings:
\textit{R1 only} to evaluate pure test-time scaling,
\textit{V3 + GEPA} as a non-reasoning model harness baseline and
\textit{V3 + \sys} as our enhanced neurosymbolic reasoning engine.
Table~\ref{tab:tte} gives a same-family, end-to-end comparison on the evaluated tool-calling tasks:
V3 with \sys exceeds R1 by 34.0 points in Avg$_1$ and vanilla V3 by 17.7.
This is not a controlled test of where reasoning should live: R1 and V3 differ in instruction following and tool-call formatting, which may partly explain R1's low scores, and the \sys pipeline also includes five fine-tuned 12B models.
It shows only that, in this setting, a contract-checked pipeline with a non-reasoning responder outperforms R1.
% We note that this does not isolate reasoning capability from other differences introduced
% by R1's post-training, including instruction following and tool-call formatting,
% and does not establish superiority on reasoning tasks outside this evaluation.

\begin{table}[t]
\centering
\footnotesize
\caption{DeepSeek-V3 with our system vs.\ R1 (reasoning, derived from V3),
vanilla V3, and GEPA. $\tau^2$ is averaged over its three domains;
Avg$_2$ uses the human-reviewed v4 score.}
\label{tab:tte}
\setlength{\tabcolsep}{3pt}
\begin{tabular}{@{}p{2.3cm} p{1.6cm} cccc@{}}
\toprule
\textbf{Method} & \textbf{Base} & \textbf{$\tau^2$} & \textbf{BFCL v4} & \textbf{Avg$_1$} & \textbf{Avg$_2$} \\
\midrule
Ours & DS-V3 & \textbf{83.5} & \textbf{70.5} & \textbf{72.4} & \textbf{75.2} \\
\midrule
Test-time scaling & DS-R1 & 33.2 & 57.0 & 38.4 & 41.5 \\
Vanilla & DS-V3 & 66.6 & 46.5 & 54.7 & 57.2 \\
GEPA & DS-V3 & 42.7 & 26.5 & 35.7 & 38.2 \\
\bottomrule
\end{tabular}
\end{table}

% \paragraph{Resource contrast.}
% The resource asymmetry between test-time scaling and neurosymbolic programs at the \emph{post-training} stage is stark. DeepSeek-R1's reasoning capability was developed through a four-stage post-training pipeline on top of the DeepSeek-V3 base model~\citep{deepseek-r1}: (1)~cold-start SFT on curated chain-of-thought examples, (2)~reasoning-focused RL using GRPO with rule-based rewards, sampling 16 outputs per question at up to 32{,}768 tokens each, (3)~rejection sampling producing $\sim$600K reasoning samples plus $\sim$200K general samples for a second SFT stage ($\sim$800K examples total), and (4)~a second RL phase for all-scenario alignment. The RL stages alone used 512 NVIDIA H800 GPUs at a reported cost of \$294K for the first stage~\citep{deepseek-r1}; the full four-stage post-training pipeline likely exceeds \$500K--\$1M when accounting for both RL stages and the SFT computation on 800K examples.
The two approaches also differ sharply in what they require at the \emph{post-training} stage. DeepSeek-R1's reasoning capability was developed through a four-stage pipeline on top of DeepSeek-V3, with two large-scale RL stages and SFT on roughly 800K examples~\citep{deepseek-r1}.
Our post-training is a single SFT stage: five SLM-based primitives fine-tuned on a 12B base model from 16{,}502 examples.
\sys instead enforces contracts and step-level checks around a non-reasoning responder, and the same SLM layer is used across responders without retraining.
\section{Discussion}
\label{sec:discussion}

Decomposition alone is not sufficient for reliable agentic reasoning; the benefit comes from coupling it with contracts, prerequisite checks, and recovery, while learned primitives handle the semantic judgments that cannot be formalized.

\paragraph{Limitations.}
Our guarantees are structural: a learned primitive can produce an output that
satisfies its contract and is still semantically wrong, and we do not yet
measure how often this occurs. Contracts check the reasoning trace, not the base
model's final response, which may use or discard parts of the trace; we do not
measure how often it departs from it. Most configurations are single runs, so small
differences between systems (for example, under one point of aggregate
accuracy) should not be read as meaningful; repeated runs with uncertainty
estimates are needed. Our ablation removes the contract-and-control layer as a
whole and does not apportion credit among contracts, prerequisite checking, and
recovery; nor do we yet compare fine-tuned primitives against prompted ones
inside the same program, which would separate the contribution of the
architecture from that of fine-tuning. \sys rows include five fine-tuned 12B models that vanilla rows lack, so comparisons are system-versus-model, not compute-matched. We do not report per-task tokens,
latency, or inference cost. Building the kernel required 300 manually worked
scenarios, and we have not yet tested transfer to unseen tool registries or
quantified the authoring effort for a new domain. Finally, our evidence is
limited to agentic tool-calling; whether other reasoning domains admit
decomposition into contract-checked primitives remains open.

\section{Conclusion}
\label{sec:conclusion}

We introduce \sys, a neurosymbolic reasoning system that composes typed
symbolic and neural primitives through an embedded DSL into \emph{reasoning
programs} whose every step is governed by an output contract enforced at
execution. On $\tau^2$-Bench, BFCL v3 and v4, and LiveMCPBench, \sys achieves
the highest aggregate accuracy among the evaluated methods (by a margin within
single-run noise over the best frontier baseline), lifts systems built on small
open models to frontier-level accuracy, and lets a pipeline with a non-reasoning
responder outperform its reasoning sibling by 34 points with a single
fine-tuning stage. Removing
the contract-and-control layer costs up to 38.6 points: structure helps only
when it is checked. For structured tool-calling, externalizing reasoning into
contract-checked programs is thus a promising alternative to scaling implicit
reasoning inside model weights; establishing which further domains admit such
decomposition is the natural next step.

\clearpage
\bibliography{refs}

\clearpage

\appendix
% Appendix - Execution Kernel Algorithm
% Wraps the full kernel algorithm in an appendix section so \ref{app:algorithm} resolves.
% Requires \usepackage{algorithm} and \usepackage{algorithmic} in the preamble.

\section{Execution Kernel Algorithm}
\label{app:algorithm}

Algorithm~\ref{alg:kernel} gives the complete control flow of the tool-calling
execution kernel (Section~\ref{sec:tool-calling}) across all five sub-processes.
SLM-based primitives are marked \textit{[SLM]}; all other operations are
deterministic. It is the same kernel as Algorithm~\ref{alg:kernel_small}, with
the branches that report missing functions and parameters made explicit. Invariant tags (INV-S, INV-T, INV-E, INV-L, INV-R) reference the
cross-process invariants defined in Section~\ref{sec:tool-calling}.

% Tool-Calling Execution Kernel algorithm, colored by primitive type.
% Requires in preamble:
%   \usepackage{algorithm}
%   \usepackage{algorithmic}
%   \usepackage{xcolor}
% SLM-based operations in blue, deterministic operations in teal/green.
% Colors chosen to stay distinguishable in grayscale; [SLM]/[det] tags are
% retained so the distinction also survives black-and-white printing.

\newcommand{\slm}[1]{#1}
\newcommand{\detp}[1]{#1}
\newcommand{\slmtag}{\textit{[SLM]}}
\newcommand{\dettag}{\textit{[det]}}

\begin{algorithm}[!ht]
\caption{Tool-Calling Execution Kernel with error-reporting branches.
\textit{[SLM]} marks SLM-based primitives and \textit{[det]} deterministic ones.
Invariant tags reference Section~\ref{sec:tool-calling}.}
\label{alg:kernel}
\begin{algorithmic}[1]
\REQUIRE input $\iota$, config $\gamma$, registry $\Omega$, turn context $\tau$
\STATE \textbf{// SP1: Decompose}
\STATE $\mathit{turns} \leftarrow \slm{\textsc{Turns}(\iota)}$ \hfill \slmtag
\STATE $\mathit{config} \leftarrow \detp{\textsc{Extract}(\gamma)}$;\ \ $\mathit{tools} \leftarrow \detp{\textsc{Register}(\Omega)}$ \hfill \dettag
\STATE \textbf{// SP2: Init-Scratchpad}
\STATE $S \leftarrow \emptyset$;\ \ \textbf{for} $(k,v) \in \mathit{config}$: $S \leftarrow S \cup \{k \mapsto v\}$ \COMMENT{INV-S}
\STATE \textbf{// SP3: Get-Order}
\STATE $G \leftarrow \slm{\textsc{BuildDepGraph}(\mathit{turns})}$ \hfill \slmtag
\STATE \textbf{if} $\detp{\textsc{CycleDetect}(G)}$ \textbf{then} halt with error \COMMENT{INV-T} \hfill \dettag
\STATE $T \leftarrow \detp{\textsc{TopoSort}(G)}$ \COMMENT{leaves first, root last} \hfill \dettag
\FORALL{$t \in T$}
    \STATE \textbf{// SP4: Check-Prerequisites}$(t, S)$
    \FORALL{$p \in \detp{\textsc{GetPrereqs}(t)}$}
        \STATE $r \leftarrow \detp{\textsc{Check}(p, S)}$ \COMMENT{$\mathit{Met}$ / $\neg\mathit{Met}$ / $\mathit{Absent}$} \hfill \dettag
        \IF{$r = \mathit{Absent}$}
            \STATE $v \leftarrow \textsc{Execute}(\slm{\textsc{TaskToCheck}(p)})$ \hfill \slmtag
            \STATE $S \leftarrow S \cup \{p \mapsto v\}$ \COMMENT{fires once per $p$ (INV-T)}
            \STATE $r \leftarrow \detp{\textsc{Check}(p, S)}$ \COMMENT{re-check retrieved value}
        \ENDIF
        \IF{$r = \neg\mathit{Met}$}
            \STATE $t' \leftarrow \slm{\textsc{TaskToMeet}(p)}$ \hfill \slmtag
            \STATE \textbf{report} $\mathit{Blocked}(t')$; \textbf{skip} to next $t$ \COMMENT{abandon $t$ only (INV-R)}
        \ENDIF
    \ENDFOR
    \STATE \textbf{// SP5: Match-Funcs-and-Params}$(t, S, \tau, \Omega)$
    \STATE $f \leftarrow \slm{\textsc{MatchFunc}(t, \Omega)}$ \hfill \slmtag
    \IF{$f = \bot$}
        \STATE \textbf{report} \textsc{MissingFunction}; \textbf{ continue} \COMMENT{abandon $t$ (INV-R)}
    \ENDIF
    \STATE $M \leftarrow \detp{\textsc{Params}(f)}$;\ \ $B \leftarrow \emptyset$ \hfill \dettag
    \FORALL{$m \in M$}
        \STATE $v \leftarrow \detp{\textsc{Lookup}(m, \tau, S)}$ \COMMENT{$\tau(m)$, else $S(m)$, else $\bot$ (INV-L)} \hfill \dettag
        \IF{$v = \bot$}
            \STATE $v \leftarrow \textsc{Execute}(\slm{\textsc{TaskToGet}(m)})$ \hfill \slmtag
            \STATE $S \leftarrow S \cup \{m \mapsto v\}$ \COMMENT{write even if $\bot$ (INV-S)}
            \STATE \textbf{if} $v = \bot$ \textbf{then report} \textsc{MissingParam}($m$); \textbf{ break}
        \ENDIF
        \STATE $B \leftarrow B \cup \{m \mapsto v\}$
    \ENDFOR
    \STATE \textbf{if} $\mathrm{dom}(B) = M$ \textbf{then} $\detp{\textsc{Execute}}(f, B)$ \COMMENT{execution guard (INV-E)} \hfill \dettag
\ENDFOR
\end{algorithmic}
\end{algorithm}

\FloatBarrier
\section{Engine Optimizations and DSL Refinement Details}
\label{app:design-details}

\paragraph{Runtime optimizations.}
Before and during execution, the engine dispatches independent calls
concurrently, caches results of previously executed primitives with identical
inputs, batches SLM calls to the same model for throughput, and prunes branches
rendered unreachable. These optimizations do not change program semantics; we
do not evaluate their effect on latency in this paper.

\paragraph{Revisions driven by the worked solutions.}
Each of the 300 manually worked solutions that revealed friction, ambiguity, or
failure in a sub-process specification led to a concrete revision of the DSL:
\begin{itemize}
    \item Cases where prerequisite checking produced false negatives identified
    missing scratchpad propagation rules, leading to the $\mathit{Absent}$
    branch with inline retrieval.
    \item Cases where parameter resolution failed identified the need for the
    prioritized lookup function with explicit turn-context-over-scratchpad
    ordering.
    \item Cases with ambiguous task ordering led to the explicit dependency
    graph construction with cycle detection.
    \item Cases where the scratchpad lost cross-turn state motivated the
    monotonicity invariant: keys are never deleted, preventing accidental loss
    of cross-turn state.
\end{itemize}

\section{Worked Design-Time Verification Solutions}
\label{app:worked-solutions}

This appendix presents representative worked solutions from the corpus of 300
manually verified DSL executions used for design-time verification
(see Methodology section). We organize them by the DSL capability each
exercises rather than by scenario, so that each example illustrates a distinct
aspect of the execution kernel. Each solution traces the five sub-processes and
compares the resulting tool-call sequence against the reference call sequence.

\subsection{Cross-Turn State Management}
\label{app:cross-turn}

Turn~2 of the scenario in Figure~\ref{fig:cross-turn} requires searching a file that a prior
turn moved into a new directory. The directory and file location are resolved
from scratchpad state established in Turn~1, while the search pattern is taken
from the current turn's user input. This exercises the scratchpad monotonicity
invariant (INV-S) and the turn-context-over-scratchpad lookup priority.
(Presented in full as Figure~\ref{fig:cross-turn}.)

\subsection{Multi-Step Dependency Chains}
\label{app:multi-step}

Turn~3 of a second file-system scenario decomposes a single request (``copy
\texttt{release\_notes.txt} into \texttt{handoff} and rename the copy'') into
three ordered unit tasks whose execution order follows the dependency chain.

\begin{lstlisting}[basicstyle=\scriptsize\ttfamily, frame=single, columns=fullflexible]
Scenario: file system, Turn 3
Input: "Copy release_notes.txt into handoff and rename the
        copy to client_release_notes.txt."

SP1 Decompose -> unit tasks:
  t1: Copy release_notes.txt to handoff
  t2: Navigate to handoff
  t3: Rename release_notes.txt -> client_release_notes.txt

SP3 Get-Order -> T = [t1, t2, t3]   (dependency chain)

SP4/SP5 (per task):
  1. cp(source='release_notes.txt', destination='handoff')
  2. cd(folder='handoff')
  3. mv(source='release_notes.txt',
        destination='client_release_notes.txt')

Reference: identical (3/3 calls)      Result: EXACT MATCH
\end{lstlisting}

\subsection{Prerequisite Checking Against Persistent State}
\label{app:prereq}

A single-task turn (``Find the nearest airport to Miami'') illustrates
prerequisite evaluation. The task carries a prerequisite
\texttt{user\_authenticated}, which the three-valued \textsc{Check} function
resolves to \textit{Met} directly from persistent scratchpad state (the
\texttt{authenticated} flag set in a prior turn), so no remediation is needed.

\begin{lstlisting}[basicstyle=\scriptsize\ttfamily, frame=single, columns=fullflexible]
Scenario: TravelAPI multi-turn, Turn 1
Input: "Find the nearest airport to Miami."
Scratchpad (carried): authenticated = true, ...

SP4 Check-Prerequisites(t1, S):
  P = { user_authenticated }
  Check(user_authenticated, S):
      S(authenticated) = true -> Satisfied -> Met
  => <Ready, S>

SP5 Match-Funcs-and-Params(t1, S, tau, Omega):
  f = get_nearest_airport_by_city   [SLM match]
  M = { city }
  Lookup(city, tau, S): tau(city) = "Miami"
  B = { city -> "Miami" },  dom(B) = M   [guard satisfied]
  => Execute(get_nearest_airport_by_city, {city: "Miami"})

Result: SUCCESS (returns MIA, Miami International Airport)
\end{lstlisting}

\subsection{Cross-Task Parameter Resolution}
\label{app:cross-task}

A passenger name correction decomposes into a retrieval task followed by an
update task. The update draws the corrected name from the turn context $\tau$
while preserving the date of birth from the scratchpad $S$ populated by the
retrieval task, a value the user did not mention and must not alter. This is
the clearest example of the lookup priority ($\tau$ before $S$) combined
with preservation of unmentioned fields. An additional variant, reservation
\texttt{NZK0WL}, follows the identical pattern.

\begin{lstlisting}[basicstyle=\scriptsize\ttfamily, frame=single, columns=fullflexible]
Scenario: Passenger name correction
Input: "Correct passenger name on reservation G2LTC4 to James Nguyen."
Known context: existing passenger Ethan Nguyen (user id ethan_nguyen_7360)

SP1 Decompose -> unit tasks:
  t1: Retrieve reservation details
  t2: Update passenger name
  t2 depends on t1 (needs existing DOB)  =>  T = [t1, t2]

Execute t1 -- Retrieve reservation:
  Check-Prerequisites(t1, S): P = {} => <Ready, S>
  f = get_reservation_details,  M = { reservation_id }
  Lookup(reservation_id, tau, S): tau -> 'G2LTC4'
  Execute -> updates S:
    S <- S + { current_passengers -> [{ dob: '1998-08-02', ... }] }

Execute t2 -- Update passenger name:
  Check-Prerequisites(t2, S):
    P = { reservation_loaded, passenger_dob_known }
    both Met from S after t1  =>  <Ready, S>
  f = update_reservation_passengers
  M = { reservation_id, passengers }
  Parameter resolution (lookup priority: tau before S):
    reservation_id : tau -> 'G2LTC4'      (user input)
    first_name     : tau -> 'James'       (user-requested change)
    last_name      : tau -> 'Nguyen'      (user-requested)
    dob            : S   -> '1998-08-02'  (preserved from t1;
                                           name correction != DOB change)

DSL output:
  1. get_reservation_details(reservation_id='G2LTC4')
  2. update_reservation_passengers(reservation_id='G2LTC4',
       passengers=[{first_name:'James', last_name:'Nguyen',
                    dob:'1998-08-02'}])

Result: SUCCESS   DOB preserved via S(t1)   name fields resolved from tau
\end{lstlisting}

% \subsection{Error-Handling Branches}

% TODO: Add worked traces for the non-success branches of the kernel once
% available. These are the highest-value additions to this appendix, as they
% demonstrate the error-catching behavior that distinguishes design-time
% verification from post-hoc checking. Target one trace for each:
%
%   (a) A prerequisite that resolves to NOT-Met, triggering a remediation
%       task (TaskToMeet) that the caller must execute before retrying t.
%
%   (b) An Absent prerequisite that triggers inline retrieval (TaskToCheck),
%       writes the retrieved value to S, and re-enters the check loop
%       (firing at most once per prerequisite key, per INV-2).
%
%   (c) A parameter that cannot be resolved from tau or S and whose retrieval
%       (TaskToGet) returns bottom, causing a MissingParam failure that
%       abandons only task t and continues with the next task (INV-R).
%
% \begin{lstlisting}[basicstyle=\scriptsize\ttfamily, frame=single]
% ...
% \end{lstlisting}

% \emph{[Worked error-handling traces to be added; see source comments.]}
\section{BFCL v4 Scoring Corrections}
\label{app:bfcl-corrections}

During evaluation we observed that the standard BFCL v4 multi-turn scoring
pipeline penalizes a recurring class of outputs that are functionally correct.
To quantify this, we performed a human review of failing cases for \emph{each
model and technique on each benchmark}; the human-reviewed scores reported in the Results
section are the result of applying this review uniformly
across all evaluated systems. The review was done by three expert annotators (about 89
annotator-hours in total). We do not enumerate per-system correction counts, we
do not claim the annotators were independent of system development, and the
corrected labels have not been independently audited. We illustrate here the \emph{kinds} of scoring issues the review
identified, with representative examples. In every case we hold a deliberately
conservative standard, described below, and we retain genuine model errors as
failures.

We treat an output as a \emph{scoring artifact} only when it reaches the same
scored state by a valid alternative path, is penalized only on surface
formatting, or performs setup the reference omits. An output that performs an
unrequested action with genuine side effects on API state is treated as a real
error, even when the final user-visible result appears similar, and is not
reclassified as an artifact. This is consistent with the paper's thesis that
correct intermediate state, not only final output, is what matters.

\subsection{Category A: Reference Omits Required Setup Calls}

Here the reference hard-codes values or assumes state that the task in fact
requires the agent to establish. Our system performs the necessary
call (resolving a city to an airport code, or authenticating before a
booking) and is penalized for output that is, if anything, more complete than
the reference.

\paragraph{Example 1.}
\emph{Request:} plan a business-class flight from ``Los Angeles'' to ``New
York'' on 2026-04-15, then book it with the given card and access token. \\
\emph{Reference:} get\_flight\_cost (with travel\_from=LAX, travel\_to=JFK),
then book\_flight. \\
\emph{Our output:} two get\_nearest\_airport\_by\_city calls resolving ``Los
Angeles'' and ``New York'' to LAX and JFK, then get\_flight\_cost,
authenticate\_travel, and book\_flight. \\
\emph{Why this is an artifact:} the user supplies city names, not airport codes.
Resolving them via get\_nearest\_airport\_by\_city is a required and correct
step; the reference silently hard-codes the codes and omits the authentication
the booking API requires. Our output is penalized for performing setup the task
genuinely demands.

\paragraph{Example 2.}
\emph{Request:} book a first-class flight from ``San Francisco'' to ``Los
Angeles'' on 2026-10-15 with the given travel card. \\
\emph{Reference:} get\_flight\_cost (with travel\_from=SFO, travel\_to=LAX),
then book\_flight. \\
\emph{Our output:} two get\_nearest\_airport\_by\_city calls resolving ``San
Francisco'' and ``Los Angeles'' to codes, then authenticate\_travel,
get\_flight\_cost, and book\_flight. \\
\emph{Why this is an artifact:} identical to the above: city-to-airport
resolution and authentication are necessary and correct, but the reference
omits them and scores their presence as divergence.

\subsection{Category B: Decomposed Calls Reaching Equivalent State}

The output performs the same effective action as the reference but reaches it
through additional or split calls that do not alter the scored end state.

\paragraph{Example 3.}
\emph{Request:} draft a tweet naming the two files, comma-separated, with the
hashtag \#fileshowcase. \\
\emph{Reference:} post\_tweet with content ``file1.txt, file2.txt'' and tag
\#fileshowcase. \\
\emph{Our output:} a read-only posting\_get\_login\_status call, then post\_tweet
with the same file names and hashtag. \\
\emph{Why this is an artifact:} the posted tweet conveys the same file names and
hashtag as the reference; the additional call is a read-only status check that
does not mutate tweet state, and the decomposition into multiple calls reaches
an equivalent end state. The exact-match scorer nonetheless flags the trace as
divergent.

\subsection{Category C: Valid Alternative Path to Equivalent State}

The output obtains the same information or reaches the same state as the
reference through a different but valid call sequence, and is penalized by exact
execution-state matching rather than by any functional difference.

\paragraph{Example 4.}
\emph{Request:} explore ResearchDocs for files or subdirectories containing the
keyword ``draft''. \\
\emph{Reference:} find with path=ResearchDocs and name=draft. \\
\emph{Our output:} ls, then cd into ResearchDocs, then find with path=. and
name=draft. \\
\emph{Why this is an artifact:} navigating into ResearchDocs and searching the
current directory covers exactly the same search space as searching ResearchDocs
directly; the two produce the same matches.

\paragraph{Example 5.}
\emph{Request:} track down a recently lodged support ticket. \\
\emph{Reference:} get\_ticket with ticket\_id=1. \\
\emph{Our output:} get\_user\_tickets. \\
\emph{Why this is an artifact:} both retrieve the target ticket; get\_user\_tickets
returns the user's tickets (including the one requested), obtaining equivalent
information through a valid alternative accessor with no difference in resulting
state.

\paragraph{Summary.}
The categories above recur across models and techniques: because the standard
pipeline scores exact call sequences and intermediate state, it systematically
penalizes correct outputs that resolve underspecified inputs, differ only in
formatting, or reach the target state by a valid alternative route. Applying the
human review uniformly across all evaluated systems yields the corrected scores
in Section~\ref{sec:results}. The correction raises every system's absolute
numbers and changes a small number of pairwise orderings (for example, Haiku~4.5
with \sys{} trails vanilla Opus~4.7 under automated scoring but leads it under
human review), which is why we report both and treat the automated scores as
primary.
\section{SLM Fine-Tuning Details}
\label{app:finetuning-examples}

Each SLM-based primitive is fine-tuned on examples that pair a task-specific
system prompt with a structured JSON target. We show one representative example
per primitive below. System prompts are abbreviated to their operative
constraints, and the model's output is shown as the structured object it must
produce; the exact-match validation criterion (Section~\ref{sec:finetuning})
requires this object to match the target. These examples are drawn from our
synthetic corpus and are illustrative of the input--output contract each
primitive learns.

\subsection{Decompose}

\emph{System (abbrev.):} break the request into atomic unit tasks, each calling
exactly one registry tool; identify dependencies; ground every parameter in the
user request, environment config, or prior conversation. \\
\emph{Input:} ``Play Marvin Gaye for 15 minutes, and after that finishes, play
Stevie Wonder for 12 minutes.'' \\
\emph{Target:}
\begin{quote}\ttfamily\footnotesize\raggedright
\{"tasks": [\{"id":"t1", "tool\_name":"spotify.play",
"depends\_on":[], "parameters":\{"artist":"Marvin Gaye", "duration":15\}\},
\{"id":"t2", "tool\_name":"spotify.play", "depends\_on":["t1"],
"parameters":\{"artist":"Stevie Wonder", "duration":12\}\}],
"config\_keys":\{\}\}
\end{quote}
The second task is marked dependent on the first, capturing the ``after that
finishes'' ordering constraint.

\subsection{Get-Order}

\emph{System (abbrev.):} given a batch of mutually independent tasks, output a
total ordering that best fits the user's workflow; output exactly the given
ids, inventing no dependencies. \\
\emph{Input:} three ready tasks: \texttt{s5} enable roaming, \texttt{s6} add
3\,GB data, \texttt{s7} toggle airplane mode. \\
\emph{Target:} \texttt{\{"ordered\_task\_ids": ["s5", "s6", "s7"]\}}

\subsection{Check-Prerequisites}

\emph{System (abbrev.):} for a prerequisite key that is \emph{absent} (missing
from scratchpad) or \emph{not\_met} (present but predicate false), propose at
most one unit task that would advance the check toward \emph{Met}. \\
\emph{Input:} key \texttt{destination\_set}, outcome \texttt{absent}; main task
``Navigate to New York after the prerequisite is resolved.'' \\
\emph{Target:}
\begin{quote}\ttfamily\footnotesize\raggedright
\{"task": \{"tool\_name":"set\_navigation",
"parameters":\{"destination":"New York"\}\}\}
\end{quote}
The absent key triggers a single setup task that will populate it.

\subsection{Match-Func}

\emph{System (abbrev.):} given one unit task and a tool registry, choose at most
one tool whose name matches the registry exactly; return null if none applies. \\
\emph{Input:} task ``Add 3GB of mobile data to my line L-555'', registry
\{\texttt{enable\_roaming}, \texttt{get\_customer\_by\_phone},
\texttt{refuel\_data}, \texttt{run\_speed\_test}\}. \\
\emph{Target:} \texttt{\{"tool\_name": "refuel\_data"\}}

\subsection{Match-Fill}

\emph{System (abbrev.):} propose values only for missing required parameters,
grounded in the task text, turn context, or scratchpad; if a value must come
from state, read it from the scratchpad rather than inventing it. \\
\emph{Input:} missing parameter \texttt{item\_ids} for
\texttt{exchange\_delivered\_order\_items}; turn context supplies
\texttt{new\_item\_ids} = [I-100, I-101]; scratchpad \texttt{S} holds
\texttt{item\_ids} = [I-200, I-201]. \\
\emph{Target:} \texttt{\{"values": \{"item\_ids": ["I-200", "I-201"]\}\}}

\medskip
\noindent This last example illustrates the lookup discipline directly: the
required \texttt{item\_ids} is resolved from the scratchpad rather than
conflated with the similarly named \texttt{new\_item\_ids} present in the turn
context, matching the priority rule enforced by the execution kernel.

\begin{table}[t]
\centering
\small
\caption{Fine-tuning dataset sizes for SLM-based primitives. All primitives use Gemma 12B as the base model with a 90/10 train/validation split.}
\label{tab:finetuning-data}
\begin{tabular}{lrrr}
\toprule
\textbf{Primitive} & \textbf{Total} & \textbf{Train} & \textbf{Val} \\
\midrule
\textsc{Decompose}            & 6{,}910  & 6{,}218  & 692 \\
\textsc{Get-Order}            & 2{,}000  & 1{,}800  & 200 \\
\textsc{Check-Prerequisites}  & 4{,}867  & 4{,}376  & 491 \\
\textsc{Match-Func}           & 1{,}600  & 1{,}440  & 160 \\
\textsc{Match-Fill}           & 1{,}125  & 1{,}011  & 114 \\
\midrule
\textbf{Total}                & 16{,}502 & 14{,}845 & 1{,}657 \\
\bottomrule
\end{tabular}
\end{table}

\begin{table}[t]
\centering
\small
\caption{Per-primitive exact-match accuracy before and after fine-tuning, on Gemma 12B.
The average is weighted by validation-set size. Fine-tuning improves every primitive,
with the largest gains on semantically more complex tasks (\textsc{Decompose},
\textsc{Check-Prerequisites}).}
\label{tab:primitive-reliability}
\begin{tabular}{@{}lcc@{}}
\toprule
\textbf{Primitive} & \textbf{Base} & \textbf{SFT} \\
\midrule
\textsc{Decompose}            & 69.7 & \textbf{92.8} \\
\textsc{Get-Order}            & 92.0 & \textbf{100.0} \\
\textsc{Check-Prerequisites}  & 72.9 & \textbf{87.4} \\
\textsc{Match-Func}           & 90.0 & \textbf{96.9} \\
\textsc{Match-Fill}           & 86.0 & \textbf{93.0} \\
\midrule
Weighted avg.       & 76.4 & \textbf{92.5} \\
\bottomrule
\end{tabular}
\end{table}
\section{Evaluation Technical Details}
\label{app:eval-details}

\paragraph{RLM.} Recursive Language Models, an agentic framework in which the base model writes and executes code to inspect its input and intermediate state, run with fine-tuned frozen code and configurable iteration budgets ($\mathit{max\_iter} = 15$--$20$, $\mathit{max\_llm\_calls} = 40$--$50$). We evaluate with Kimi K2.6, GPT-OSS-120B, and DeepSeek-V3 as inference models. DeepSeek-V3 results appear only in Table~\ref{tab:main-results} (three of six benchmarks errored out) and are not part of the sweep below. RLM pass rates are computed on potentially reduced denominators: runs that produce errors are excluded from the denominator, yielding an effective sample size $N$ that may be smaller than the full benchmark. We report the error rate alongside scores to enable fair comparison. Table~\ref{tab:rlm-sweep} shows the sweep for Kimi K2.6 and GPT-OSS-120B; we use the best-performing configuration per benchmark in our main comparison.

A notable observation is RLM's high error rate with GPT-OSS-120B (18--34\%), indicating that the framework's reliability is sensitive to the choice of inference model. On reduced denominators, pass rates appear competitive, but when errors are counted as failures against the full benchmark size, effective accuracy drops substantially.

\begin{table}[t]
\centering
\footnotesize
\caption{RLM sweep. ft = fine-tuned frozen code. $N$ = effective sample size after excluding errors. Best pass\% per benchmark in \textbf{bold}.}
\label{tab:rlm-sweep}
\setlength{\tabcolsep}{2.5pt}
\begin{tabular}{@{}lllrl@{}}
\toprule
\textbf{Bench.} & \textbf{Params} & \textbf{Inf.} & \textbf{Pass / Rwd} & \textbf{Err} \\
\midrule
BFCL    & baseline          & kimi   & \textbf{56.0} / 59.2 & -- \\
        & i20, c50 ft       & kimi   & 50.5 / --    & -- \\
\midrule
Airline & baseline          & kimi   & 81.6 / 98    & 2\% \\
        & i20, c50 ft       & kimi   & \textbf{86.6} / 90   & 10\% \\
        & i15, c40 ft       & kimi   & 86.4 / 88    & 12\% \\
        & i20, c50 ft       & gpt    & 21.2 / 66    & 34\% \\
\midrule
Retail  & baseline          & kimi   & 73.7 / --    & -- \\
        & i20, c50 ft       & kimi   & \textbf{80.6} / 94.7 & 5\% \\
        & i15, c40 ft       & kimi   & 70.6 / 89.5  & 11\% \\
        & i20, c50          & gpt    & 34.6 / 71    & 29\% \\
\midrule
Telecom & baseline          & kimi   & 81.6 / --    & -- \\
        & i20, c50 ft       & kimi   & \textbf{79.6} / 95.6 & 4\% \\
        & i20, c50 ft       & gpt    & 54.8 / 82    & 18\% \\
\bottomrule
\end{tabular}
\end{table}

\paragraph{GEPA.} A genetic evolution-based prompt optimization framework for structured tool-calling. GEPA evolves a task-specific prompt via a task model and a reflection model within a metric-call budget ($\mathit{max\_metric\_calls} = 150$--$250$). To keep the comparison fair, GEPA was compiled on the same synthetic training set that was used to build \sys{}, not on the benchmark test sets. We observed that scores are sensitive to the compilation recipe (choice of task model, reflection model, and budget). We evaluate compiled prompts with DeepSeek V3, Kimi K2.6, MiniMax M3, and GPT-OSS-120B as inference models. Table~\ref{tab:gepa-sweep} lists the recipes we ran, with scores on our synthetic set; these are not benchmark test scores and are not comparable to Table~\ref{tab:main-results}, which reports the compiled prompts on the benchmark test sets. Not every recipe was evaluated in every domain.

\begin{table}[t]
\centering
\footnotesize
\caption{GEPA sweep, scored on our synthetic set (not the benchmark test sets of Table~\ref{tab:main-results}). Recipe = task / reflection model. Best per domain in \textbf{bold}. ``--'' = not evaluated.}
\label{tab:gepa-sweep}
\setlength{\tabcolsep}{3pt}
\begin{tabular}{@{}llll@{}}
\toprule
\textbf{Bench.} & \textbf{Recipe} & \textbf{Inf.} & \textbf{Pass / Rwd} \\
\midrule
BFCL    & ds-v3 / kimi    & ds-v3     & 43.5 / 47.1 \\
        & ds-v3 / kimi    & kimi      & 62.5 / 66.1 \\
        & kimi / kimi     & kimi      & 71.5 / 77.0 \\
        & kimi / kimi     & mm-m3     & \textbf{79.0 / 83.3} \\
\midrule
Airline & ds-v3 / kimi         & kimi    & 76.0 \\
        & gpt / mm, 150c       & gpt     & \textbf{82.0} \\
\midrule
Retail  & kimi / kimi          & kimi    & -- \\
        & gpt / mm, 250c       & gpt     & \textbf{82.6} \\
        & gpt / mm, 250c       & kimi    & 78.1 \\
        & gpt / mm, 250c       & mm-m3   & 81.6 \\
\bottomrule
\end{tabular}
\end{table}

\section{Preliminary MCP-Atlas Results}
\label{app:mcp-atlas}

MCP-Atlas is excluded from the main results (Section~\ref{sec:results}) because
we have not completed evaluation across the full set of models and
methods. We report the partial results we do have here, for
transparency and because the pattern is consistent with our main findings. These
numbers should be read as preliminary: the model coverage is incomplete, no
prompt-orchestration or test-time-scaling baselines were run, and the scores are
single-run.

\begin{table}[t]
\centering
\footnotesize
\caption{Preliminary MCP-Atlas results. Matched pairs (same base model, with and
without \sys) are the meaningful comparison; coverage is incomplete and these
scores are excluded from all aggregate figures in the main text.}
\label{tab:mcp-atlas}
\setlength{\tabcolsep}{4pt}
\begin{tabular}{@{}l l c@{}}
\toprule
\textbf{Method} & \textbf{Model} & \textbf{MCP-Atlas} \\
\midrule
\sys      & GPT-OSS-120B & 61.4 \\
\sys      & DeepSeek V3  & 57.0 \\
\midrule
Vanilla   & Sonnet 4.6   & 68.5 \\
Vanilla   & GPT-OSS-120B & 32.8 \\
Vanilla   & DeepSeek V3  & 43.4 \\
\bottomrule
\end{tabular}
\end{table}

\paragraph{Matched comparisons.}
For the two base models evaluated under both settings, \sys{} produces large
gains: GPT-OSS-120B improves from 32.8 to 61.4 ($+28.6$ points, $+87\%$
relative), and DeepSeek V3 from 43.4 to 57.0 ($+13.6$ points, $+31\%$ relative).
The ordering is consistent with the main results, and with the pattern that the
weaker base model benefits more.

\paragraph{Frontier comparison.}
Vanilla Sonnet 4.6 attains the highest absolute score on this benchmark at 68.5,
above both \sys{} configurations we were able to run. We did not evaluate \sys{}
with Sonnet 4.6 or any other frontier base model on MCP-Atlas, so this table does
not support a claim about \sys{} relative to frontier models on this benchmark in
either direction. Completing that comparison is left to future work.

\paragraph{Evaluator caveat.}
MCP-Atlas scores tasks using an LLM judge over reference claims. Recent audits of
tool-calling benchmarks report meaningful evaluator--human misalignment for
judge-based benchmarks of this kind, so single-run scores should be treated as
indicative rather than precise. This is a further reason we exclude MCP-Atlas
from the aggregate figures reported in the main text.


\end{document}